# Implicit Dual-domain Convolutional Network for Robust Color Image Compression Artifact Reduction

Bolun Zheng, Yaowu Chen, and Xiang Tian, Fan Zhou, Xuesong Liu

1`

*Abstract*—Several dual-domain convolutional neural network-based methods show outstanding performance in reducing image compression artifacts. However, they are unable to handle color images as the compression processes for gray scale and color images are different. Moreover, these methods train a specific model for each compression quality, and they require multiple models to achieve different compression qualities. To address these problems, we proposed an implicit dual-domain convolutional network (IDCN) with a pixel position labeling map and quantization tables as inputs. We proposed an extractor-corrector framework-based dual-domain correction unit (DCU) as the basic component to formulate the IDCN; the implicit dual-domain translation allows the IDCN to handle color images with discrete cosine transform (DCT)-domain priors. A flexible version of IDCN (IDCN-f) was also developed to handle a wide range of compression qualities. Experiments for both objective and subjective evaluations on benchmark datasets show that IDCN is superior to state-of-the-art methods and IDCN-f exhibits excellent abilities to handle a wide range of compression qualities with little trade-off against performance; further, it demonstrates great potential for practical applications.

B. Zheng is with the Institute of Advanced Digital Technology and Instrument, Zhejiang University, Hangzhou 310027, China (e-mail: zhengbolun1024@163.com).
Y. Chen is with the State Key Laboratory of Industrial Control Technology, Zhejiang University, and also with the Embedded System Engineering Research Center, Ministry of Education of China, Zhejiang University, Hangzhou 310027, China (e-mail: yaowuchen@zju.edu.cn).
X. Tian is with the Institute of Advanced Digital Technology and Instrument, Zhejiang University, and is also with the Zhejiang Provincial Key Laboratory for Network Multimedia Technologies, Hangzhou 310027, China (e-mail: tianx@mail.bme.zju.edu.cn).
F. Zhou is with the Institute of Advanced Digital Technology and Instrument, Zhejiang University, and is also with the Zhejiang Provincial Key Laboratory for Network Multimedia Technologies, Hangzhou 310027, China
X. Liu is with the Institute of Advanced Digital Technology and Instrument, Zhejiang University, and is also with the Zhejiang Provincial Key Laboratory for Network Multimedia Technologies, Hangzhou 310027, China (e-mail: 11015006@zju.edu.cn)

## I. INTRODUCTION

LOSSY image compression algorithms use the information redundancy of image patches to achieve a high compression ratio with desirable image qualities. The human eye is not sensitive to high-frequency information, and therefore, most lossy image compression methods are implemented by quantization or approximation on the frequency domain. However, as the compression ratio increases, the artifacts introduced by the severe degradation of high-frequency information significantly reduce the quality of visual experience. JPEG is a representative lossy compression standard and is used globally. JPEG compression standard divides the image into $8 \times 8$ patches and performs discrete cosine transform (DCT) on each patch; then, the DCT coefficients are quantized and encoded into bit streams.

Studies on artifact reduction (AR) in image compression have recently been conducted and validated based on JPEG compression standards. Traditional filter-based algorithms [1], [2] focus on general image denoising and play a role in the AR problem. Machine learning-based methods [3], [4], [5], [6] are more engaged in specific AR problems and present a promising performance in both subjective and objective evaluations. These methods learn nonlinear mapping from the compressed image to the original image. Chen et al. [4] trained nonlinear reaction diffusion models for image denoising and JPEG deblocking. Inspired by SRCNN [7], Dong et al. [5] first introduced a deep neural network (DNN) and built a four-layer full convolutional neural network (CNN) to solve the AR problem; it was called the ARCNN. Previous studies on cascaded CNNs (CAS-CNNs) [8] and dual-domain multi-scale CNNs (DMCNNs) [9] imported MCNNs and learned large-scale features to remove large-scale artifacts. Because JPEG compression artifacts are mainly caused by the lossy quantization of DCT coefficients, Liu et al. [10] proposed a dual-domain image dictionary to recover the compressed images from both the pixel domain and the DCT domain. As DCT is a linear transformation that can be easily rewritten in a convolutional pattern, Guo et al. [11] proposed a dual-domain convolutional network (DDCN) and improved its performance on gray-scale images.

Despite their high performance on gray-scale images, DCT domain-based methods typically suffer from handling color images for two main reasons. First, because the entire network is a highly nonlinear system, simple linear operations on input images would lead to unpredictable nonlinear outputs. In other words, using methods developed for gray-scale images to recover color images by restoring each color channel separately would lead to chromatic aberrations (Fig. 1). Second, because the compression algorithm for the luminance channel differs from that for the two chrominance channels, using a single model trained for the luminance channel to recover two chrominance channels would produce undesired results. However, for restoring color images compressed by JEPG, although pixel-domain learning methods can easily learn the correlations between each color channel, it is difficult to directly introduce the DCT-domain priors.

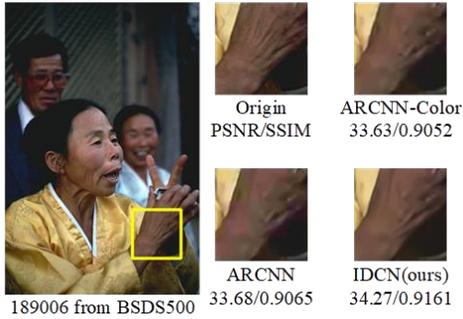

Fig. 1. Comparison between methods for gray scale and color images. Here, ARCNN-Color refers to the ARCNN trained on color images.

The mapping relationship between the compression quality factor and the quantization coefficient can be expressed as

$$\varepsilon(q) = \begin{cases} 200 - 50q & q \in \mathrm{N}^{(50,100]} \\ \dfrac{5000}{q} & q \in \mathrm{N}^{[1,50]} \end{cases}, \quad (1)$$

$$Q^q = \Re\left(\dfrac{\varepsilon Q}{100}\right), \quad (2)$$

where $q$ is the compression quality factor, $Q$ is the base quantization coefficient table, $Q^q$ is the quantization coefficient table with quality factor $q$, and $\Re$ is the rounding operation. When $q$ reaches a relatively low range, a slight change in $q$ leads to a great change in $Q^q$. Thus, the performance of recovering DCT coefficients based on the prior of quantization coefficients severely declines when $q$ of the recovered image mismatches the $q$ of the pre-trained model.

To address these problems, we propose a method to reduce high-quality color image compression artifacts by developing an implicit dual-domain convolutional network (IDCN) to implicitly utilize both pixel-domain features and DCT-domain priors. The main contributions of our work are summarized as follows:

1) We propose a unified architecture called IDCN to learn dual-domain corrections for the reduction in color image compression artifacts. To overcome the limitations of DCT, IDCN directly estimates the DCT-domain losses without DCT to build a color-to-color network.
2) We propose an extractor–corrector framework to construct generalized residual connection and develop the dual-domain correction unit (DCU) based on this framework.
3) We introduce dilated convolution layers in the extractor of DCU to enlarge the receptive field, which shows great performance to remove wide-range distortions.
4) We propose a position labeling map to label the position of each pixel in the entire image and concatenate the position labeling map to the input image so that the network can utilize the position information to refine the compressed image.
5) We propose a flexible IDCN (IDCN-f) for robust color image compression AR. IDCN-f exhibits great performance in handling a wide range of compression qualities with one model weight.

## II. RELATED WORK

Several CNN-based methods present promising potential to solve the image compression AR problem. In this section, we briefly review and discuss the methods relevant to this work.

Because the human eye is more sensitive to luminance than chrominance, early CNN-based methods focus on recovering the luminance channel of a JPEG compressed color image or only a gray-scale image. Dong et al. [5] first proposed a deep learning based method (ARCNN) for image compression AR. ARCNN is composed of four convolution layers: functions of these layers can be defined as feature extraction, feature enhancement, mapping, and reconstruction. However, as the network goes deeper, the function of each convolution layer is increasingly fused. Guo et al. [11] introduced a dual-domain learning architecture and built a 30-convolutional-layer network (DDCN) to utilize both the pixel-domain features and DCT-domain priors. DDCN consists of three branches: DCT-domain, pixel-domain, and aggregation, each containing 10 convolution layers. The DCT-domain branch replaces DCT and inverse DCT (iDCT) by equivalent convolution operations; quantization rectification is accomplished before the iDCT operation. Although dense DCT and iDCT convolution can utilize more redundant information from the surrounding image patches to correct the DCT coefficients of the 8 × 8 image patches, it may also mislead the corrections for those middle image patches between the adjacent 8 × 8 image patches. Inspired by the success of CNN for single-image super resolution (SR), Cavigelli et al. [8] translated the AR problem to the SR problem and introduced stepped convolution and deconvolution [12] layers (also known as up-sampling layers) to build a cascaded convolutional network (CAS-CNN). CAS-CNN introduced a multi-scale loss function to restore the compressed image in a different scale. Multi-scale loss can enlarge the gradients for the bottom layers to update the learnable weights and learn large-scale features to recover the details. However, because the proportion of larger-scale feature maps is too low, the benefits from the multi-scale learning are limited. Noticing the merits of DDCN and CAS-CNN, Zhang et al. [9] proposed a DMCNN. They adopted the dual-domain framework and multi-scale loss and introduced dilated convolution layers to enlarge the receptive fields for the removal of banding effects. Not only is the DCT domain considered to formulate the dual domain learning network, wavelet domain is also introduced to build a dual domain architecture. Chen et al. [13] proposed a dual pixel-wavelet domain deep CNN (DPW-SDNet) for soft decoding of JPEG-compressed images. DPW-SDNet made down-sampling and discrete wavelet transformation (DWT) on compressed image to get the input of pixel domain branch, and wavelet domain branch, and then, they synthesized the output of two branches to obtain a soft decoded image. Liu et al. [14] combined the multi-scale method and DWT, proposing a multi-level wavelet CNN (MWCNN) for general image restoration. MWCNN used DWT to generate multi-scale feature maps, and it used inversed DWT (IWT) to up-sample the feature maps back to the origin scale. Recently, Zheng et al. [15] identified the importance of the correlation of each color channel and constructed a scalable convolutional network (S-Net) to learn a nonlinear mapping for color JPEG compressed image restoration. Some CNN based image restoration methods also include the JPEG compression AR. Zhang et al. [16] proposed a DnCNN to handle image denoising problem, which is also effective to handle SR and compression AR problems. Zhang et al. [17] combined residual connection and dense connection, and proposed a residual dense network (RDN) for image restoration. Mao et al. [18] introduced a series of symmetric skip connected encoder–decoder pairs, and they proposed RED-Net to restore noisy images. Tai et al. [19] introduced short path, long path transmission, and gate unit, and they proposed a persistent memory network (MemNet) for image restoration.

Huang et al. [20] proposed DenseNet to ensure maximum information flow between any layers within the same block and achieved state-of-the-art performance in high-level computer vision tasks. Following DenseNet, several low-level computer vision algorithms such as SRDenseNet [21], MemNet[19], DCPDN [22], and RDN[17] introduced dense connections and achieved impressive performances in respective fields.

All these methods for AR showed significant improvement over conventional filter-based methods. However, none of them could effectively process the color images with DCT-domain priors. To address this problem, we propose an IDCN to represent the DCT-domain priors in the pixel-domain and fuse the corrections from both the DCT domain and the pixel domain.

## III. IMPLICIT DUAL-DOMAIN CONVOLUTION NETWORK

### A. Implicit Translation from DCT-domain to Pixel-domain

A major obstacle to applying DCT-domain-based methods to color image processing is that the compressing operation for luminance channels is completely different from that for chrominance channels. First, chrominance channels should be downsampled by a factor of 2 before DCT, whereas luminance channels do not need downsampling. Second, the quantization table for luminance channels is different from that for chrominance channels. Conventional DCT-domain-based methods directly recover the DCT coefficients, which is difficult to implement in color images.

JPEG compression works on YCbCr mode images. The relationship between the YCbCr color space and the RGB color space can be expressed as

$$\begin{cases} R = 1.164(Y-16) + 1.596(Cr-128) \\ G = 1.164(Y-16) - 0.392(Cb-128) - 0.813(Cr-128) \\ B = 1.164*(Y-16) + 2.017(Cb-128) \end{cases}, \quad (3)$$

where $Y$, $Cb$, and $Cr$ denote the luminance channel and two chrominance channels, respectively, and $R$, $G$, and $B$ denote the red, green, and blue channels, respectively. Therefore, we can easily calculate the RGB loss $\delta_{R,G,B}$ caused by JPEG compression

$$\delta_{R,G,B} = (\frac{\partial R}{\partial Y}, \frac{\partial G}{\partial Y}, \frac{\partial B}{\partial Y})\delta_Y + (\frac{\partial R}{\partial Cb}, \frac{\partial G}{\partial Cb}, \frac{\partial B}{\partial Cb})\delta_{Cb} \\ + (\frac{\partial R}{\partial Cr}, \frac{\partial G}{\partial Cr}, \frac{\partial B}{\partial Cr})\delta_{Cr}, \quad (4)$$

where $\delta_Y$, $\delta_{Cb}$, and $\delta_{Cr}$ denote the $Y$ channel loss, $Cb$ channel loss, and $Cr$ channel loss caused by JPEG compression, respectively.

The losses caused by JPEG compression are from the quantization operations on DCT coefficients. Considering $\Theta$ as an $8 \times 8$ DCT coefficient matrix and $\Theta^*$ as the quantized result of $\Theta$, we can define $\Theta$ as

$$\Theta = \Theta^* + \delta_\Theta, \quad (5)$$

where $\delta_\Theta$ denotes the quantization loss. $\delta_\Theta$ can be expressed as

$$\delta_\Theta = \vartheta * Q^q, \quad (6)$$

where $*$ denotes the element-wise multiplication and $\vartheta$ is the relative quantization loss, which is an $8 \times 8$ matrix and satisfies

$$-0.5 < \vartheta_i < 0.5 \quad \forall \vartheta_i \in \vartheta. \quad (7)$$

For an $8 \times 8$ image patch, the channel loss $\delta_s$, $s \in \{Y, Cb, Cr\}$ in the YCbCr color space caused by the quantization of DCT coefficients can be written as

$$\begin{aligned}\delta_{s_{i,j}} &= \kappa(\Theta)_{i,j} - \kappa(\Theta^*)_{i,j} = \kappa(\delta_\Theta)_{i,j}, i,j \in N^{[0,7]} \\ &= \kappa(\vartheta * Q^q)_{i,j} \\ &= \sum_{u=0}^{7}\sum_{v=0}^{7}\alpha(u)\alpha(v)\vartheta_{u,v}Q_{u,v}^q f(i,j,u,v) \\ &= \vartheta * \xi(Q^q, i, j)\end{aligned}, \quad (8)$$

$$\begin{cases}\xi(Q^q,i,j)_{u,v} = \alpha(u)\alpha(v)Q_{u,v}^q f(i,j,u,v) \quad u,v \in N^{[0,7]} \\ \alpha(u) = \sqrt{\frac{1}{8}} \quad\quad\quad\quad\quad\quad\quad\quad\quad u = 0 \\ \alpha(u) = \sqrt{\frac{2}{8}} \quad\quad\quad\quad\quad\quad\quad\quad\quad u > 0\end{cases}, \quad (9)$$

$$f(i,j,u,v) = \cos\frac{(2i+1)u\pi}{16}\cos\frac{(2j+1)v\pi}{16}, \quad (10)$$

where $i$ and $j$ denote the horizontal and vertical coordinates in the $8\times 8$ image patch and $\kappa$ denotes the iDCT operation. Where $\delta_s$, $\vartheta$, and $Q^q$ are reshaped into a $1\times 1\times 64$ vector, (8) can be rewritten as

$$\delta_s = \vartheta * \xi^*_{1\times 1}(Q^q), \quad (11)$$

$$\xi^*_{1\times 1}(Q^q)_{8j+i} = \xi(Q^q)_{i,j}, \quad (12)$$

where $\xi^*$ is a $1\times 1\times 64\times 64$ 2D convolution kernel. Therefore, we can use a 2D convolution to translate the DCT-domain loss to the pixel-domain loss in the YCbCr color space so that the element-wise operation in (6) can be avoided; this will reduce the calculations by approximately 40%. If we ignore the loss from the downsampling operation on chrominance channels, $\delta_Y$, $\delta_{Cb}$, and $\delta_{Cr}$ in (4) can be rewritten as

$$\delta_\chi = \vartheta_\chi * \xi^*(Q^q_\chi), \chi \in \{Y, Cb, Cr\}. \quad (13)$$

The quantization table of the $Cb$ channel is the same as that of $Cr$; thus, $Q^q_{Cr} = Q^q_{Cb} = Q^q_C$.

Estimating $\vartheta_Y$, $\vartheta_{Cb}$, and $\vartheta_{Cr}$ from the compressed image is a highly nonlinear task, which can always be handled in the pixel-feature domain and the estimated losses calculated from $\vartheta_Y$, $\vartheta_{Cb}$, and $\vartheta_{Cr}$ are in the YCbCr color space. Although we can translate the losses into the RGB color space by simple linear operations, we would rather translate the estimated loss back to the pixel-feature domain for two main reasons. First, the estimation task is handled in the feature domain; this translation can maintain the integrity of the whole nonlinear operation so that we can easily introduce skip connections to construct a deeper network. Second, for a specific pixel location in the image, the estimated $\delta_\chi$ is a $1\times 1\times 64$ vector and it represents an $8\times 8$ image patch, i.e., the $\delta_\chi$ of the locations in the same $8\times 8$ stepped patch would be close to each other and the blocking effect probably exists in $\delta_\chi$. This translation can help avoid the block effect of the deviation of $\delta_\chi$. Because the entire translation is operated from the pixel-feature domain to the pixel-feature domain and utilizes the prior from the DCT-domain, we call this translation an implicit dual-domain translation.

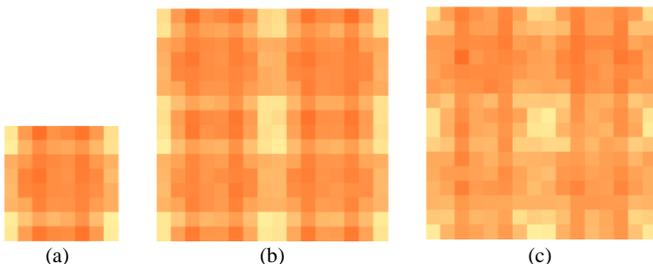

Fig. 2. Standard deviation maps with $q = 20$. (a) Patch size is 8, WIN143. (b) Patch size is 16, WIN143. (c) Patch size is 16, 200 testing images of BSDS500.

### B. Position Labeling

Because most elements in the quantization table are different from each other and the loss of DCT coefficients at different positions lead to different losses in the RGB color space, the RGB color space loss caused by the quantization of different pixels in an $8\times 8$ patch would probably be different from each other. We conducted an experiment on the DIV2K dataset to validate our assumption. As the chrominance channels are downsampled before DCT and quantization operations, we separated the luminance channel from the image and sliced the luminance channel into $8\times 8$ patches in accordance with the JPEG image. Then, we calculated the standard deviation of the difference between the original image and the JPEG compressed image at different positions. The result is shown in Fig. 2(a). The standard deviations of the elements around the corners are larger than in the middle. To further support this conclusion, we increased the patch size to 16 and kept the moving step unchanged. The standard deviation map (Fig. 2(b)) varies in cycles of $8\times 8$, which is the same as the field of the DCT patch. We also conducted this experiment on other public image datasets (Fig. 2(c)). Although the ranges of standard deviation are different, the regularity of distributions is close. If we expand this result to the entire RGB color space, the minimum period of the standard deviation should be $16\times 16$.

Based on this conclusion, it is necessary to consider the position of the pixels. The deviation is introduced by quantization in the DCT-domain. Based on the randomness of the quantization operation, we can use a zero-mean Gaussian distribution to simply describe this deviation. Thus, the deviation distribution at a specific position could be defined as

$$\psi_{i,j} = \mathbb{N}(0, \sigma_{i,j}) \quad i, j \in \mathbb{N}^{[0,7]}, \quad (14)$$

where $\mathbb{N}$ denotes the Gaussian distribution and $\sigma_{i,j}$ denotes the standard deviation at position $(i, j)$. Inspired by FFDNet [23], we used the standard deviation to label the pixel position and we called this standard deviation map a "labeling map." The labeling map has the same size as the input image, and it is defined as

$$L^\chi(x, y) = \sigma^\chi_{\text{mod}(x,\omega), \text{mod}(y,\omega)}, \chi \in \{R, G, B\}, \quad (15)$$

where $L$ denotes the labeling map and $\omega$ denotes the minimum period of standard deviation in each axis. For color images, the channel size of the labeling map is 3 and $\omega = 16$ (Fig. 3).

However, in practice, the above multi-channel labeling map did not work very well for training a single quality network. Too much unchanged labeling information may mislead the network to learn too much features from the labeling map rather than the image itself. To overcome the problem, we simplified the labeling map by

$$L(x, y) = \sqrt{L^R(x, y)^2 + L^G(x, y)^2 + L^B(x, y)^2}, \quad (16)$$

where $L$ denotes the simplified labeling map. We used this simplified labeling map for training single-quality networks,

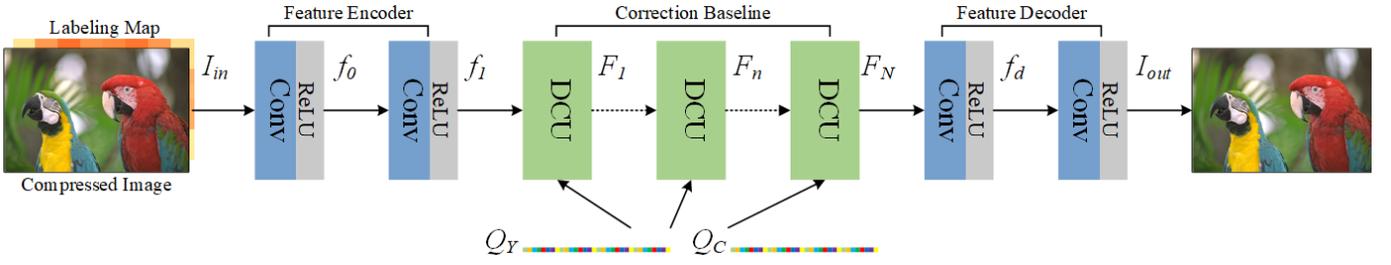

Fig. 4. Architecture of our proposed implicit dual-domain convolutional network (IDCN)

and the multi-channel labeling map for training a flexible network.

### C. Architecture

Fig. 4 illustrates the architecture of IDCN, which consists of three parts: feature encoder (FE), correction baseline (CB), and feature decoder (FD). We denote $I_{in}$ and $I_{out}$ as the input and output of IDCN. Specifically, based on the previous studies, the batch-normalization [24] layer was not considered. FE takes two convolution layers to encode the input from the pixel domain to the pixel-feature domain and each convolution layer is linked to a ReLU [25] layer. The first $5 \times 5$ convolution layer and the ReLU layer extract features $f_0$ from $I_{in}$.

$$f_0 = \text{Max}(\mathbb{C}_{FE1}(I_{in}),0), \quad (17)$$

where $\mathbb{C}_{FE1}$ denotes the convolution operation of the first convolution layer of FE. The second $3 \times 3$ convolution layer and the ReLU layer then further enhance $f_0$ to obtain $f_1$ for feature-domain corrections.

$$f_1 = \text{Max}(\mathbb{C}_{FE2}(f_0),0), \quad (18)$$

where $\mathbb{C}_{FE2}$ denotes the convolution operation of the second convolution layer of FE. $f_1$ is then used as an input to CB, which consists of a series of DCUs. Assuming that CB contains N DCUs, the output $F_n$ of the $n$-th DCU can be expressed as

$$F_n = \mho_{CB,n}(F_{n-1}) = \mho_{CB,n}(\mho_{CB,n-1}(...(\mho_{CB,1}(f_1))...)), \quad (19)$$

where $\mho_{CB,n}$ denotes the dual-domain correction operation of the $n$-th DCU. The details on DCU are provided in Section III.D. The corrected feature ($F_N$) output by CB is then sent to FD. The FD takes two convolution layers to translate $F_N$ to the final output $I_{out}$. The first $3 \times 3$ convolution layer and the ReLU layer preprocess the corrected features ($F_N$) to decodable features ($f_d$).

$$f_d = \text{Max}(\mathbb{C}_{FD1}(F_N),0), \quad (20)$$

where $\mathbb{C}_{FD1}$ denotes the convolution operation of the first convolution layer in FD. The second $5 \times 5$ convolution layer then translates $f_d$ to $I_{out}$.

$$I_{out} = \mathbb{C}_{FD2}(f_d), \quad (21)$$

where $\mathbb{C}_{FD2}$ denotes the convolution operation of the second convolution layer in FD. In our IDCN, $\mathbb{C}_{FE1}$, $\mathbb{C}_{FE2}$, and $\mathbb{C}_{FD1}$ have the same number of filters and $\mathbb{C}_{FD2}$ have three filters, as we output the color images.

### D. Dual-domain Correction Unit

The DCU is designed based on residual learning and on the results presented in Section III.A. We compared several versions of residual connections (Fig. 5). Fig. 5 (a) shows the original residual connection in ResNet [26]. SRResNet [27] (Fig. 5(b)) modified the original residual connection by removing the outside ReLU layer for single-image super resolution. The enhanced deep super-resolution network (EDSR) [28] (Fig. 5(c)) further improved the residual connection in SRResNet by removing the batch-normalization layers and concatenating the scaling layer to the residual branch to avoid gradient explosion. The residual branch can be regarded as an extractor–corrector framework. For example, in Fig. 5(c), the first convolution layer and the ReLU layer can be

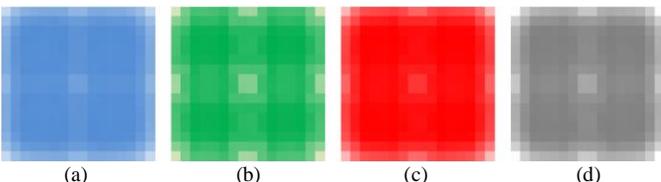

(a) (b) (c) (d)
Fig. 3. Standard deviation maps with $q = 20$ on DIV2K dataset. (a) B channel. (b) G channel. (c) R channel. (d) Simplified labeling map.

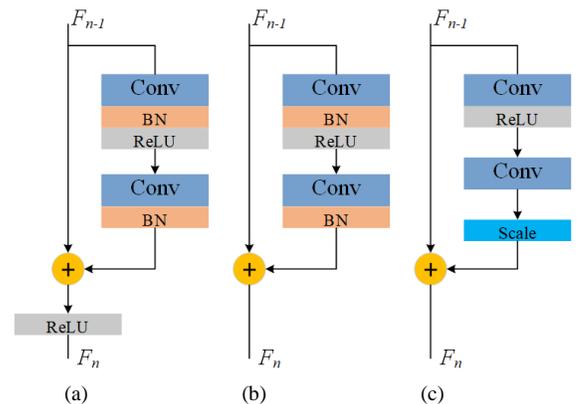

Fig. 5. Comparison of residual connections in original ResNet, SRResNet, and EDSR. (a) Original ResNet. (b) SRResNet. (c) EDSR.

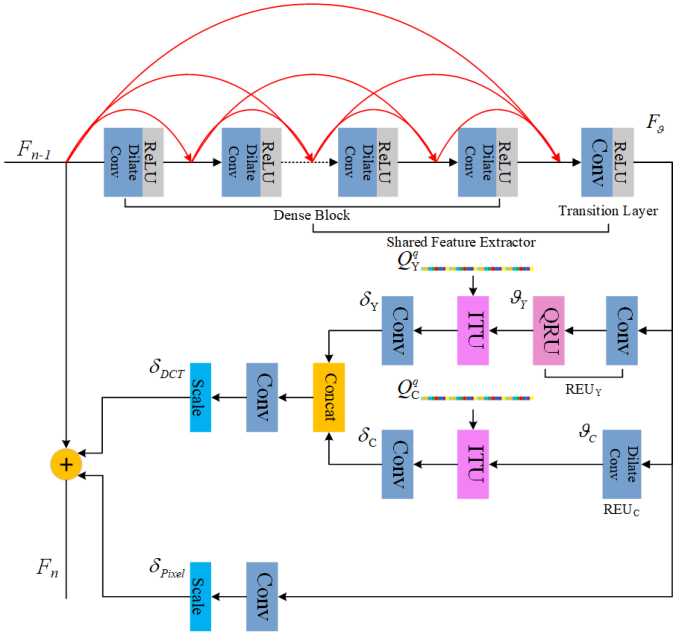

Fig. 6. Architecture of our proposed dual-domain correction unit.

regarded as a simple feature extractor and the second convolution layer and the scaling layer as a simple corrector.

In our DCU, we reconstructed the feature extractor by importing the dense block and the transition block. We also reconstructed the corrector by introducing dual correction branches, DCT-domain and pixel-domain, so that our DCU can make the corrections through both pixel and DCT domains. The structure of DCU is shown in Fig. 6. The two correction branches share the same feature extractor, shared feature extractor (SFE). Let $F_{n-1}$ and $F_n$ be the input and output of the $n$-th DCU. SFE takes $F_{n-1}$ as the input and outputs the basic features $F_g$ for the DCT-domain branch to estimate DCT-domain correction $\delta_{DCT}$ and pixel-domain correction $\delta_{Pixel}$. Then, we used a residual connection to connect $F_{n-1}$, $\delta_{DCT}$, and $\delta_{Pixel}$ together to obtain $F_n$ as:

$$F_n = F_{n-1} + \delta_{DCT} + \delta_{Pixel}. \quad (22)$$

SFE consists of a dense block and a transition block. The dense block is a group of densely connected 3 × 3 convolution layers to extract high-dimensional features. The transition block is a 1 × 1 convolution layer with the ReLU activation layer to compress the high-dimensional features to low-dimensional features. We assumed that SFE contains $L$ convolution layers and the growth is $K$. The number of filters of the first convolution layer in the dense block was set to 2$K$. Then, the number of filters of the remaining $L$-2 convolution layers was set to $K$. The outputs of the first $L$-1 convolution layers were concatenated together to be sent to the transition block. Moreover, we adopted the receptive field theory in DMCNN and introduced dilated convolution [29] to enlarge the receptive field of feature extractors. For the first L-1 densely connected layers, the first half of the layers are normal convolution layers, (dilation rate is 1) and then, the dilation rate is increased by 1 for each stacked convolution layer.

In the DCT-domain branch, relative quantization loss estimating unit (REU) takes $F_g$ as input and uses a 3 × 3 convolution layer and quantization rectified unit (QRU) to estimate the relative quantization loss $\vartheta$ in a specific channel of $Y$, $Cr$, and $Cb$. Here, QRU is a simple constraint layer to let $\vartheta$ satisfy (7). One REU estimates the loss from a channel of $Y$, $Cb$, and $Cr$. However, taking three REUs to estimate $\vartheta_Y$, $\vartheta_{Cb}$, and $\vartheta_{Cr}$ would introduce too many learnable parameters, which would increase the computation consumption and make the network difficult to converge. Because the $Y$ channel is a major channel for the human visual sense and the $Cb$ and $Cr$ channels share the same quantization table, we use one REU, $REU_Y$, to estimate $\vartheta_Y$ and another one, $REU_C$, to estimate the integrated chrominance loss $\vartheta_C$. Because $REU_C$ estimates the integrated loss from both Cr and $Cb$ channels, the constraints based on (7) do not work on $\vartheta_C$; thus, QRU is not included in $REU_C$. Moreover, because the two chrominance channels are compressed in the DCT domain after the 2 × 2 downsampling operation, we introduced a 3 × 3 dilated convolution layer with a dilation rate of 2 for $REU_C$, whereas $REU_Y$ is a normal 3 × 3 convolution layer. Then, the implicit translation units (ITUs) generated by $Q_Y^q$ and $Q_C^q$ translate $\vartheta_Y$ and $\vartheta_C$ to the luminance loss $\delta_Y$ and chrominance loss $\delta_C$, respectively, based on (8–13). Finally, we concatenated $\delta_Y$ and $\delta_C$ together and used a 3 × 3 convolution layer to fuse them to obtain the final pixel-feature domain correction of DCT-domain branch $\delta_{DCT}$. In the pixel-domain branch, we used a 3 × 3 convolution layer as a simple corrector to obtain the pixel-feature domain correction of pixel-domain branch $\delta_{Pixel}$. To avoid gradient explosion, we introduced a scale layer in EDSR to scale both $\delta_{DCT}$ and $\delta_{Pixel}$. The scaling rate was set to 0.1.

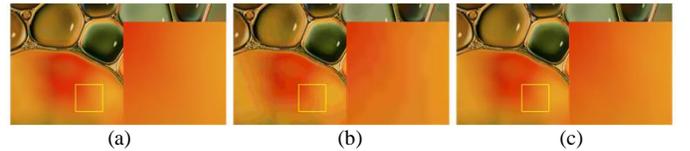

(a)        (b)        (c)
Fig. 7. 041 in WIN143 recovered by models with –DL option and –ND option. (a) Origin. (b) –ND. (c) –DL.

## IV. EXPERIMENTAL EVALUATION

### A. Implementation Details

**Network Settings**. In the proposed IDCN, CB has $N = 8$ DCUs and each DCU has the same structure, $K = 64$ and $L = 8$. Except the second convolution layer in FD, all other convolution layers in FE and FD and all trainable convolution layers in two correction branches have B = 64 filters. In particular, the number of filters of the convolution layers in REU is fixed to 64 because the relative quantization losses of 64 DCT coefficients are output.

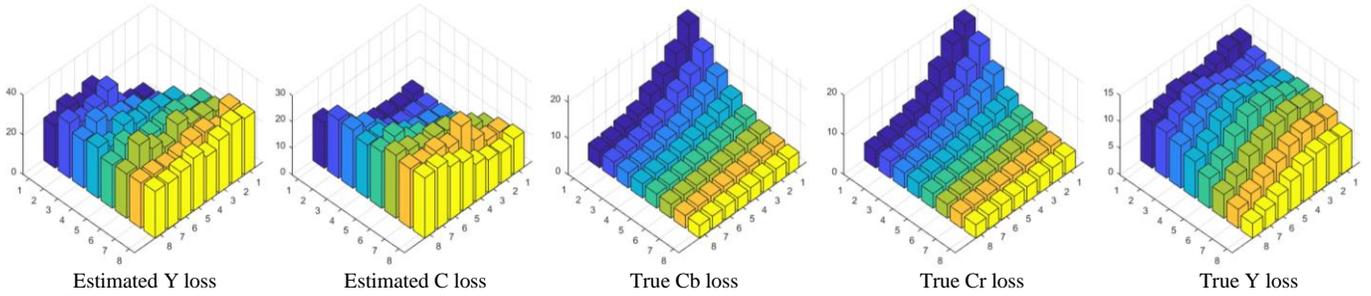

| Estimated Y loss | Estimated C loss | True Cb loss | True Cr loss | True Y loss |

Fig. 8. Comparison between estimated DCT domain loss and true DCT domain loss.

**Datasets and Metrics**. The DIV2K [30] dataset has been recently released for single-image SR. DIV2K consists of 1000 high-quality (2K resolution) images (800 training images, 100 validation images, and 100 test images). Because the test images were prepared for single-image SR competition and the ground truth has not been released, these images were not included in our experiment. We trained our models on 900 images including 800 training images and 100 validation images and conducted validation and testing on other three standard datasets: BSDS500 [31], LIVE1 [32], and WIN143 [15]. We used half of 100 validation images of BSDS500 for validation during training. For testing, quantitative evaluations were conducted on 200 test images of BSDS500 (B200), 29 images of LIVE1, and 143 images of WIN143.

For quantitative evaluations, we used standard MATLAB library functions. The objective performance indicators peak signal-to-noise ratio (PSNR), structural similarity index (SSIM) and PSNR-B were measured. Unlike most previous methods only considering the performance on the luminance channel, we measured PSNR on the full color image, and SSIM and PSNR-B are obtained by calculating the mean values of the results of the R, G, and B channels.

**Training Settings**. All model weights of the proposed IDCN were trained using the same training settings. We regarded the mean square error (MSE) loss as the loss function

$$MSE(\hat{Y}, Y) = \frac{1}{N}\sum_{i=1}^{N}(\hat{Y}_i - Y)^2, \quad (23)$$

where $Y$ is the ground truth, $\hat{Y}$ is the output of network, and $N$ is the batch size. We used Adam [33] as our training optimizer with reference parameters. The learning rate was set to $10^{-4}$ for all layers. If the decrease in the validation loss is lower than 0.001 dB for five consecutive epochs, the learning rate is divided by 5. If the learning rate is lower than $10^{-6}$, it is set to $10^{-6}$ and kept unchanged. If the learning rate reaches $10^{-6}$ and the decrease in the validation loss is lower than 0.001 dB for five consecutive epochs, the training is stopped. Referring to the settings in DMCNN, we randomly cropped 43 × 43 RGB patches with the batch size of 16 as the input and then enlarged the patch size to 96 × 96. The batch size is cut to 8 when the learning rate reaches $10^{-6}$. For patches input with a size of 43 × 43, 5000 batches constitute an epoch and for those with a size of 96 × 96, 3000 batches constitute an epoch. We implemented the IDCN using the Keras framework with Tensorflow backend. Training IDCN takes approximately seven days on a K80 GPU server for 100 epochs.

TABLE II
INVESTIGATION OF DUAL-DOMAIN LEARNING.

|  | PSNR(dB) | SSIM | PSNR-B (dB) |
|---|---|---|---|
| IDCN-P | 29.57 | 0.8741 | 29.53 |
| IDCN-Y | 29.64 | 0.8752 | 29.60 |
| IDCN-D | 29.69 | 0.8759 | 29.66 |

*B. Ablation Investigation*

In this section, we investigate the effects of three optional items: position labeling (PL), multi-channel correction (MCC), and dilated convolution (DC). PL has three options: no position labeling (-N), simplified labeling synthesized by (16) (-SL), and multi-channel labeling synthesized by (15) (-ML). MCC has two options: full correction (-FC), and simple correction (-SC). DC has two options: constructing extractors with dilated convolutions (-DL), and without dilated convolution (-ND). Here, full correction refers to making corrections from Y, Cb, and Cr, and simple correction refers to making corrections from the luminance channel and the combined chrominance channel.

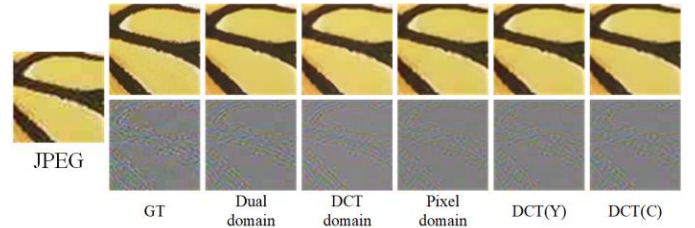

Fig. 9. Correction maps from dual domain, DCT domain, pixel domain, luminance channel and combined chrominance channel in DCT domain in IDCN-D.

Table I lists the results of the ablation investigation of the effects of options on PL and MCC. Four networks have the same network settings; $N = 3$, $K = 20$, $L = 6$, and $B = 64$. This investigation was conducted on LIVE1 with $q = 20$ and the performance indicator is PSNR.

Using the -FC option may lead to worse results. Too many subbranches in the DCT-domain correction branch cannot further improve the performance but may scatter the loss backpropagation. Using the -SC option instead can effectively simplify the network and make a significant improvement. Selecting the -SL option can also improve the performance by introducing simplified pixel position information to avoid misleading the training. Even though using the -DL option leads to little performance enhancement, enlarging the receptive of extractor could effectively help remove the wide range artifacts as shown in Fig. 7.

TABLE I
ABLATION INVESTIGATION OF PL, MCC AND DC OPTIONS.

| DC Option | | PL Option | | | MCC Options | | PSNR (dB) |
|---|---|---|---|---|---|---|---|
| -ND | -DL | -N | -SL | -ML | -FC | -SC | |
|  | √ | √ |  |  | √ |  | 29.51 |
|  | √ |  |  | √ | √ |  | 29.55 |
|  | √ | √ |  |  |  | √ | 29.59 |
| √ |  |  |  | √ |  | √ | 29.66 |
|  | √ |  |  | √ |  | √ | 29.69 |
|  | √ |  | √ |  |  | √ | 29.73 |

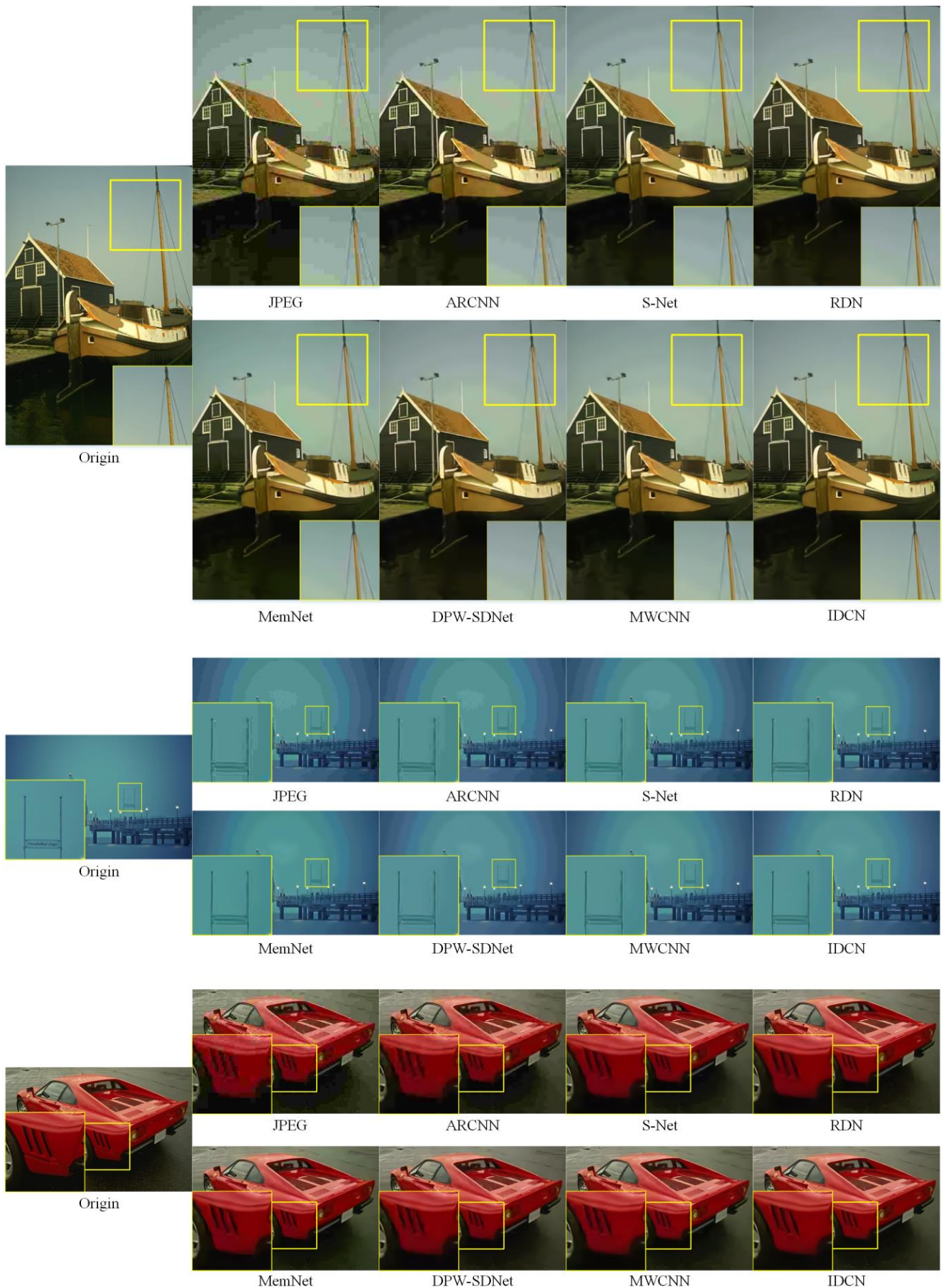

Fig. 10. Subjective evaluation with compression quality *q*=10. From top to bottom: 140088 from B200, 005 from WIN143 and 29030 from B200.

### C. Investigation of Dual-domain Learning

**Analysis on network architecture**. To investigate the effects of dual-domain learning, we developed three networks which are learned only in the pixel domain (IDCN-P), the pixel domain and only luminance channel in the DCT domain (IDCN-Y), and both the pixel domain and the DCT domain (IDCN-D). These three networks have the same settings as described in Section IV.B, and all of them have the same options, which are -SL, -SC and -D. Table II provides the experimental results obtained on the LIVE1 dataset. It is evident that IDCN-Y is a little better than IDCN-P, and IDCN-D achieves the best performance on all PSNR and SSIM PSNR-B indicators.

**Analysis on DCU**. We investigated the effects of different subbranches in DCU. Fig. 9 shows the correction maps from IDCN-D, which are calculated in the pixel domain, the DCT domain, the luminance channel in the DCT domain, and the combined chrominance channel in the DCT domain. The correction maps calculated in the DCT domain could output more smooth results, while several high-frequency artifacts remain in the pixel domain output. The correction maps calculated by the luminance channel and the combined chrominance channel are relatively closed such that the difference only exists around the edge positions. This phenomenon is caused by two reasons. First, the luminance channel and the combined chrominance channel have very closed structures, and therefore, they receive closed gradients in the training procedure, which finally lead to closed outputs. Second, to match the downsampling operation in the compressing procedure, the dilated convolution is introduced for estimating the loss in the combined chrominance channel, which will probably lead to a low-frequency output.

TABLE III
ARCHITECTURE DETAILS OF DIFFERENT METHODS.

| Method | Layers | Parameters |
|---|---|---|
| ARCNN | 5 | 117K |
| DWP-SDNet | 40 | 1.3M |
| MemNet-M6R6 | 80 | 2.8M |
| MWCNN | 24 | 15.4M |
| RDN(8 RDBs) | 85 | 13.2M |
| S-Net | 20 | 10.2M |
| IDCN | 100 | 9.9M |

**Analysis on DCT-domain learning**. We measured the mean norms of DCT coefficients of different frequencies by

$$E(fx, fy) = \sqrt{Mean((\sum_k \delta_{s_{fx,fy}}^k)^2)} \quad (24)$$

where $fx$ and $fy$ denote the horizontal and vertical coordinates in DCT coefficient map and $\delta_{s_{fx,fy}}^k$ denotes the DCT coefficient loss estimated by the $k$-th DCU. We visualized the mean norms of the estimated DCT coefficient losses of the luminance channel and the combined chrominance channel, and we compared them with the true DCT coefficient loss of the luminance channel and the two chrominance channels. The visualized results are shown in Fig. 8. We have three observations: (1) The estimated DCT coefficient losses are focused on a high-frequency area, which well matches the observation in the DCU analysis that the DCT domain could output more smooth results. (2) Because all the operations on estimated DCT-domain loss are linear, considering the effect from the scale layer, the mean norms of the estimated DCT coefficient losses are statistically closed to the true losses around the high-frequency area. (3) The mean norms of the estimated DCT coefficient losses are around zero, which is caused by the implicit transformation itself, in that the mean absolute values of implicit transformation coefficients ($\xi(Q^q)$) of low frequencies are much smaller than those of high frequencies. In general, the DCT-domain learning could statistically estimate the high-frequency components of DCT coefficient losses.

TABLE IV
QUANTITATIVE COMPARISON FOR HANDLING RGB IMAGES ON LIVE1, B200 AND WIN143. AVERAGE **PSNR/SSIM** /**PSNR-B** VALUES FOR COMPRESSION QUALITY $q$=10 AND 20. THE BEST RESULTS ARE **HIGHLIGHTED** AND THE SECOND BEST RESULTS ARE UNDERLINED.

| Quality | Methods | LIVE1 | | | B200 | | | WIN143 | | |
|---|---|---|---|---|---|---|---|---|---|---|
| | | PSNR | SSIM | PSNR-B | PSNR | SSIM | PSNR-B | PSNR | SSIM | PSNR-B |
| 20 | JPEG | 28.06 | 0.8409 | 27.82 | 28.20 | 0.8483 | 27.90 | 29.47 | 0.8440 | 29.28 |
| | ARCNN | 29.23 | 0.8659 | 29.24 | 29.36 | 0.8665 | 29.34 | 30.82 | 0.8776 | 30.92 |
| | DPW-SDNet | 29.59 | 0.8744 | 29.55 | 29.67 | 0.8752 | 29.62 | 31.28 | 0.8866 | 31.31 |
| | MemNet | 29.76 | 0.8770 | 29.75 | 29.80 | 0.8776 | 29.78 | 31.47 | 0.8904 | 31.56 |
| | MWCNN | 29.80 | 0.8769 | 29.78 | 29.85 | 0.8789 | 29.82 | 31.55 | 0.8916 | 31.63 |
| | RDN | 29.84 | 0.8778 | 29.82 | 29.85 | 0.8779 | 29.83 | 31.54 | 0.8912 | 31.63 |
| | S-Net | 29.81 | 0.8781 | 29.79 | 29.86 | 0.8782 | 29.82 | 31.47 | 0.8904 | 31.56 |
| | IDCN | **30.04** | **0.8816** | **30.01** | **30.07** | **0.8816** | **30.02** | **31.82** | **0.8964** | **31.90** |
| 10 | JPEG | 25.69 | 0.7592 | 0.25.49 | 25.83 | 0.7584 | 25.58 | 27.08 | 0.7684 | 26.90 |
| | ARCNN | 26.91 | 0.7946 | 26.92 | 27.02 | 0.7930 | 27.02 | 28.46 | 0.8207 | 28.57 |
| | DPW-SDNet | 27.26 | 0.8036 | 27.28 | 27.39 | 0.8027 | 27.39 | 29.03 | 0.8326 | 29.13 |
| | MemNet | 27.33 | 0.8100 | 27.34 | 27.46 | 0.8086 | 27.46 | 29.04 | 0.8380 | 29.15 |
| | MWCNN | 27.45 | 0.8083 | 27.44 | 27.52 | 0.8069 | 27.52 | 29.25 | 0.8375 | 29.34 |
| | RDN | 27.47 | 0.8116 | 27.48 | 27.53 | 0.8096 | 27.53 | 29.19 | 0.8395 | 29.30 |
| | S-Net | 27.35 | 0.8090 | 27.36 | 27.42 | 0.8066 | 27.43 | 28.95 | 0.8349 | 29.05 |
| | IDCN | **27.63** | **0.8161** | **27.63** | **27.69** | **0.8136** | **27.69** | **29.45** | **0.8467** | **29.56** |

TABLE V
QUANTITATIVE COMPARISON FOR HANDLING LUMINANCE CHANNEL IMAGES ON LIVE1, B200 AND WIN143. AVERAGE **PSNR/SSIM /PSNR-B** VALUES FOR COMPRESSION QUALITY $q$=10 AND 20. THE BEST RESULTS ARE **HIGHLIGHTED** AND THE SECOND BEST RESULTS ARE <u>UNDERLINED</u>.

| Quality | Methods | LIVE1 | | | B200 | | | WIN143 | | |
|---|---|---|---|---|---|---|---|---|---|---|
| | | PSNR | SSIM | PSNR-B | PSNR | SSIM | PSNR-B | PSNR | SSIM | PSNR-B |
| 20 | JPEG | 30.07 | 0.8683 | 29.64 | 30.04 | 0.8671 | 29.53 | 32.40 | 0.8924 | 31.79 |
| | ARCNN | 31.41 | 0.8891 | 31.37 | 31.32 | 0.8872 | 31.26 | 33.80 | 0.9124 | 33.75 |
| | DPW-SDNet | 31.69 | 0.8952 | 31.60 | 31.59 | 0.8933 | 31.49 | 34.12 | 0.9175 | 34.05 |
| | MemNet | 31.82 | 0.8970 | 31.74 | 31.71 | 0.8950 | 31.62 | 34.27 | 0.9190 | 34.19 |
| | MWCNN | 31.90 | 0.8989 | 31.83 | 31.78 | 0.8966 | 31.69 | 34.41 | 0.9207 | 34.33 |
| | RDN | <u>31.93</u> | <u>0.8991</u> | <u>31.85</u> | <u>31.81</u> | <u>0.8986</u> | <u>31.71</u> | <u>34.43</u> | <u>0.9207</u> | <u>34.34</u> |
| | S-Net | 31.83 | 0.8975 | 31.76 | 31.71 | 0.8952 | 31.62 | 31.28 | 0.9192 | 34.20 |
| | IDCN | **32.09** | **0.9006** | **32.00** | **31.95** | **0.8981** | **31.86** | **34.60** | **0.9220** | **34.51** |
| 10 | JPEG | 27.77 | 0.7905 | 27.38 | 27.79 | 0.7874 | 27.33 | 29.92 | 0.8248 | 29.37 |
| | ARCNN | 29.11 | 0.8235 | 29.07 | 29.08 | 0.8209 | 29.03 | 31.44 | 0.8641 | 31.40 |
| | DPW-SDNet | 29.40 | 0.8320 | 29.34 | 39.35 | 0.8292 | 29.28 | 31.84 | 0.8717 | 31.78 |
| | MemNet | 29.44 | 0.8327 | 29.39 | 29.38 | 0.8295 | 29.32 | 31.89 | 0.8727 | 31.85 |
| | MWCNN | 29.55 | 0.8357 | 29.49 | 29.47 | 0.8325 | <u>29.42</u> | <u>32.09</u> | <u>0.8754</u> | <u>32.04</u> |
| | RDN | <u>29.56</u> | <u>0.8359</u> | <u>29.51</u> | <u>29.47</u> | <u>0.8325</u> | 29.42 | 32.05 | 0.8751 | 32.01 |
| | S-Net | 29.44 | 0.8325 | 29.39 | 29.38 | 0.8294 | 29.32 | 31.86 | 0.8718 | 31.82 |
| | IDCN | **29.71** | **0.8384** | **29.66** | **29.61** | **0.8347** | **29.55** | **32.23** | **0.8773** | **32.19** |

### D. Comparison with State-of-the-Art Methods

We compare IDCN with the state-of-the-art CNN based methods on color image compression AR. Considering that most existing methods focus on gray scale image restoration, we reproduced several representative methods including ARCNN, MemNet, RDN, MWCNN, and DWP-SDNet for color image restoration to avoid the situation presented in Fig. 1. For fairness, all these methods are reproduced with the same training settings provided in original publications and trained on DIV2K (no augmentation). Owing to hardware limitations, we reproduced RDN by reducing the number of residual dense blocks (RDB) to 8, while it is 16 in the original paper. However, because the source codes of DDCN and DMCNN are not released and their DCT domain branches cannot be extended to handle color images, these two methods were not included in the comparison. We list the architecture details of the compared methods in Table III. The comparison is conducted on an open release benchmark, including LIVE1, 200 testing images of BSDS500 (B200), and the WIN143 datasets. The quantitative results are presented in Table IV. It is clear that IDCN exhibits a significant improvement on all datasets, compression qualities and metrics compared to those of the other state-of-the-art methods. As $q$ becomes lower, the gap is further increased. We also present the qualitative results in Fig. 10. The proposed method produces more defined edges, and more accurate colors whereas the block effects, blurrings, and color aberrations more or less exist in the results of other methods. Moreover, IDCN is more robust against wide-range color distortion, whereas the color band phenomenon cannot be totally removed by other methods. In general, IDCN generates more visually pleasing results.

To further demonstrate the effectiveness and superiority of proposed method, we also compare proposed method with the relevant methods on luminance channel image compression AR. However, even though some of them have released their codes and pre-trained model weights for handling luminance channel images, these codes and weights are obtained from different computing frameworks (e.g. Caffe, Matlab, Pytorch and Tensorflow, etc.) and datasets. For fairness, we reproduced these methods by following the settings in previous color image compression AR comparison. To make the proposed method could well match the gray scale image input, we made two modifications that recalculating the labeling map of luminance channel on DIV2K dataset, and removing the combined chrominance subbranch. The qualitative results are presented in Table V. As it shown, the result of luminance component comparison is roughly the same as RGB channel comparison, that our IDCN surpasses all compared methods and achieves the state-of-the-art performance.

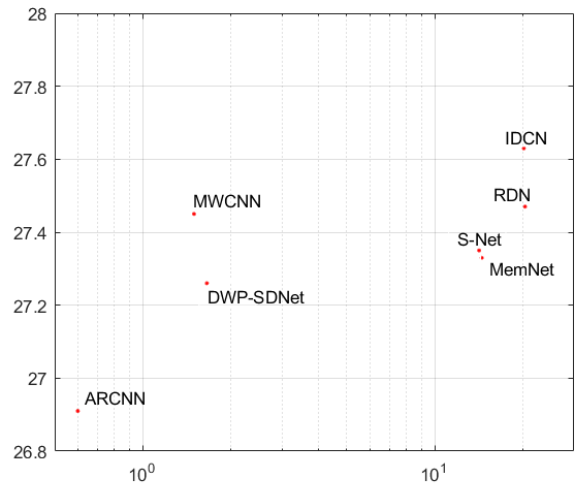

Fig. 11. Performances and running times of different methods on LIVE1 dataset with $q = 10$.

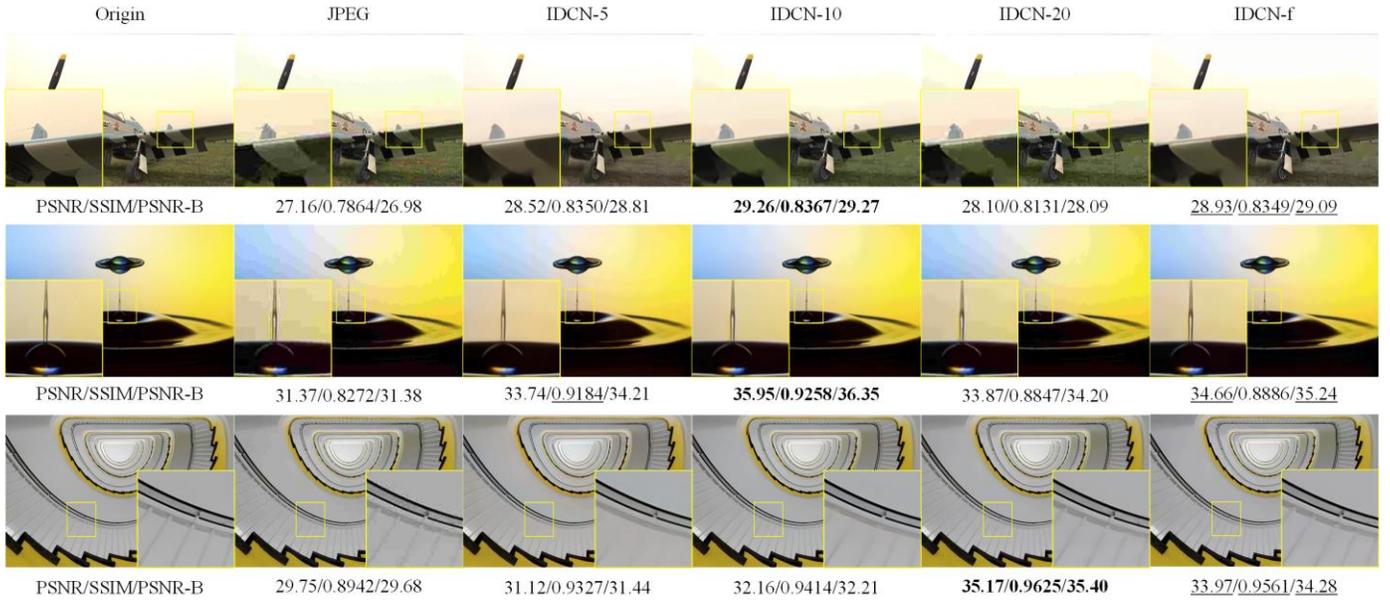

Fig. 13. Subjective evaluation for robust compression artifact reduction. From top to bottom: plane from LIVE1 and $q = 7$, 024 from WIN143 and $q = 10$, 038 from WIN143 and $q = 20$. The best results are **highlighted** and the second-best results are underlined.

Moreover, we measured the running times for the different methods. Fig. 11 illustrates the performances and running times of different methods handling the LIVE1 dataset on a 2080Ti GPU. Owing to the downsampling operation, MWCNN and DWP-SDNet received excellent computational efficiencies. ARCNN, RDN, MemNet, S-Net, and our IDCN work on the original image resolution. Our IDCN has a running speed that is very similar to the simplified RDN, and it is a bit slower than MemNet and S-Net.

### E. Robust Compression Artifact Reduction

For image compression, especially in JPEG standards, the compression quality $q$ is always known. We trained a flexible IDCN (IDCN-f) to handle the compressed images with a different $q$. Because the labeling map is generated by statistical priority for different $q$, it is not convenient to calculate all statistical priorities of different $q$ values. We assume that IDCN-f is trained for compression qualities in $N^{[ql,qh]}$. We first calculate the labeling maps $L_{ql}$ and $L_{qh}$ of $ql$ and $qh$, respectively, and the labeling map of q in $N^{[ql,qh]}$ is generated by

$$L_q = \frac{\varepsilon(q)-\varepsilon(ql)}{\varepsilon(qh)-\varepsilon(ql)}L_{qh} + \frac{\varepsilon(qh)-\varepsilon(q)}{\varepsilon(qh)-\varepsilon(ql)}L_{ql}. \quad (24)$$

To cover the low compression quality, we trained IDCN-f in $N^{[5,20]}$. The quantitative results of IDCN-f on LIVE1 are shown in Fig. 12.

IDCN-f achieved high-performance levels for all $q$ values. Although there is a slight performance degradation compared to the specifically trained IDCN models, IDCN-f produces visual pleased outputs in a wide quality range. Fig. 8 shows the performance curves of IDCN-f and other specifically trained IDCN models IDCN-5, IDCN-10 and IDCN-20 in the compression quality range of $N^{[5,20]}$. Here, IDCN-5, IDCN-10, and IDCN-20 refer to the IDCN model trained with compression qualities $q = 5$, 10, and 20, respectively. IDCN-f exhibits great robustness to handle variant compression qualities, whereas the specifically trained IDCN models cannot handle those mismatched compression qualities. Fig. 13 shows the subjective comparison between IDCN-5, IDCN-10, IDCN-20, and IDCN-f. IDCN-5 obtained terrible results when handling larger compression qualities for both objective and subjective evaluations. Although IDCN-10 could obtain relatively better results than IDCN-20 and IDCN-5, the chromatic aberration and missing details remained. In contrast, IDCN-f could fix both chromatic aberrations and texture details for a wide range of compression qualities.

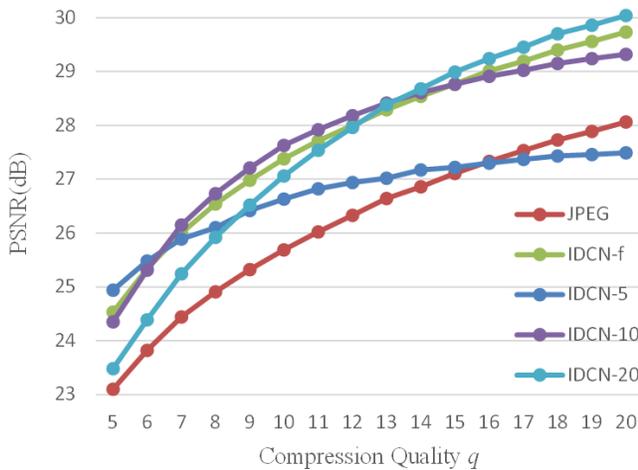

Fig. 12. Performance curves of JPEG, IDCN-f, IDCN-10, and IDCN-20. The averaged PSNR values are evaluated on LIVE1.

## V. Conclusions

In this study, we proposed a novel network based on implicit dual-domain learning (IDCN) to reduce color image compression artifacts. We first analyzed the structure of the conventional residual connections for low-level computer vision tasks and presented the extractor-corrector framework for constructing more generalized residual connections. Then, we constructed a correction unit based on this framework and introduced dense connections for the extractor and dual-domain correction for the corrector. Unlike conventional dual-domain learning methods that introduce DCT to enter the DCT-domain, the losses in the DCT-domain were directly estimated from the extracted features without DCT. The DCU also benefits the larger receptive field from the dilated convolution layers to obtain superior results in low-compression quality images. We compared IDCN with other state-of-the-art methods on both widely used evaluation datasets and the expanded WIN143 dataset. The objective and subjective evaluations demonstrated the superiority of IDCN over the other state-of-the-art methods. Moreover, the flexible version of IDCN named IDCN-f exhibited excellent performance to handle variable compression qualities in both subjective and objective evaluations. Therefore, IDCN-f has a considerable potential for applications in the practical compression AR efforts.

## VI. Acknowledgment

This work was supported in part by the Fundamental Research Funds for the Central Universities.

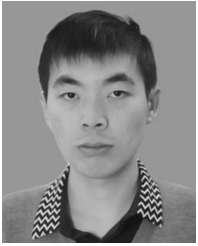
**Bolun Zheng** received the B.Sc. degree from Zhejiang University in 2014 and now pursuing the Ph.D. degree in engineering instrument science and technology with Zhejiang University, Hangzhou, China. His current research interests include image processing, video processing and deep learning.

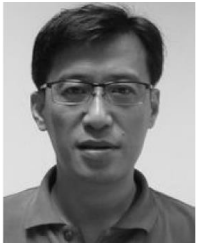
**Yaowu Chen** received the Ph.D. degree from Zhejiang University, Hangzhou, China, in 1998. He is currently a Professor and the Director of the Institute of Advanced Digital Technologies and Instrumentation, Zhejiang University. His current research interests include embedded system, multimedia system, and networking.

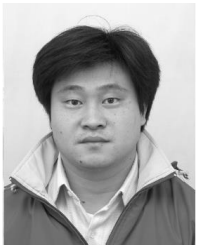
**Xiang Tian** received B.Sc. and Ph.D. degrees in signal processing from Zhejiang University, Hangzhou, China, in 2001 and 2007, respectively. He is currently an Associate Professor at Zhejiang University. His research focuses on the fields of signal processing and video coding.

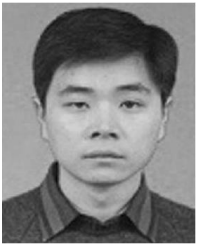
**Fan Zhou** received the B.Sc. and Ph.D. degrees from Zhejiang University, Hangzhou, China, in 2000 and 2006, respectively. He is currently an Associate Professor with Zhejiang University. His current research interests include sonar signal processing and fieldprogrammable gate array-based high-performance computing.

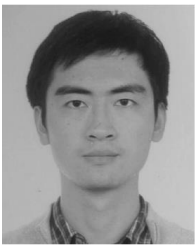
**Xuesong Liu** received the B.Sc. and Ph.D. degrees in sonar signal processing from Zhejiang University, Hangzhou, China, in 2010 and 2015, respectively. He is currently a Lecturer with Zhejiang University. His current research interests include sonar signal processing, parallel processing, and embedded system design.